\documentclass[letterpaper]{article} % DO NOT CHANGE THIS
\usepackage{aaai2027}  % DO NOT CHANGE THIS
\usepackage[hyphens]{url}  % DO NOT CHANGE THIS
\usepackage{graphicx} % DO NOT CHANGE THIS
\usepackage{natbib}  % DO NOT CHANGE THIS AND DO NOT ADD ANY OPTIONS TO IT
\usepackage{caption} % DO NOT CHANGE THIS AND DO NOT ADD ANY OPTIONS TO IT
\usepackage{amsmath}
\usepackage{amssymb}
\usepackage{bm}

\usepackage[ruled,vlined,linesnumbered,noend]{algorithm2e}

\SetCommentSty{plaincommentsty}

\usepackage{booktabs}   % \toprule, \midrule, \bottomrule
\usepackage{multirow}   % \multirow
\usepackage{colortbl}   % \rowcolor
\usepackage{tabularx}

\newcolumntype{C}{>{\centering\arraybackslash}X}
\usepackage{siunitx}
\usepackage{pifont}     % \ding

\usepackage{xcolor}
\definecolor{softred}{RGB}{255, 0, 0}
\definecolor{softgreen}{RGB}{36, 131, 11}

\usepackage{xspace}
\newcommand{\system}{\textsc{Hugin}\xspace}
\newcommand{\ie}{{\em i.e., \xspace}}

\newcommand{\eg}{{\em e.g., \xspace}}

\newcommand{\squishlist}{
	\begin{list}{$\bullet$}{
			\setlength{\itemsep}{0pt}
			\setlength{\parsep}{3pt}
			\setlength{\topsep}{3pt}
			\setlength{\partopsep}{0pt}
			\setlength{\leftmargin}{3.5mm}
			\setlength{\labelwidth}{1em}
			\setlength{\labelsep}{0.5em}}}
	\newcommand{\squishend}{\end{list}}

\title{{\system: Enhancing Vision-Language Planning for Autonomous Logistics Sorting}
\thanks{Preprint. This manuscript is currently under review.}
}
\author{
    Xikai Sun\textsuperscript{\rm 1},
    Cangtian Zhou\textsuperscript{\rm 2},
    Kebin Liu\textsuperscript{\rm 1},
    Ke Ma\textsuperscript{\rm 3},
    Xu Wang\textsuperscript{\rm 1},\\
    Zaishu Chen\textsuperscript{\rm 4},
    Haotian Wang\textsuperscript{\rm 4},
    Li Liu\textsuperscript{\rm 1},
    Yunhao Liu\textsuperscript{\rm 1}
}

\affiliations{
    \textsuperscript{\rm 1}Tsinghua University, Beijing, China    
    \textsuperscript{\rm 2}Harbin Institute of Technology, Harbin, China\\
    \textsuperscript{\rm 3}Northwestern Polytechnical University, Xi'an, China    
    \textsuperscript{\rm 4}JD Logistics, Beijing, China 
}

\begin{document}
	
	\maketitle

	\begin{abstract}
		
		Autonomous logistics sorting systems (ALSS) are an important industrial application of embodied AI, which requires joint planning over spatially disjoint camera views. We formulate this setting as Joint Multi-Scene Understanding (JMSU). With open-world visual understanding and task-planning capabilities, vision-language models (VLMs) are promising candidates for JMSU. However, directly applying existing VLMs to JMSU is non-trivial due to scarce cross-scene supervision and attention dispersion caused by long visual context in JMSU. To address these challenges, we propose \system\footnote{In Norse mythology, Hugin is one of Odin's ravens, serving as his eyes to fly across the world and gather information. Similarly, our scheme plays a similar role, helping the VLM better perceive and understand distributed scenes.}, a training framework with two complementary components. Endogenous Data Augmentation recombines verified atomic facts under operating constraints, while Global Context Ranking aligns the instruction representation more strongly with the complete visual context than with a partial visual context.  To support ongoing research, we construct a high-quality industrial sorting dataset and benchmark named SortingBench from four layouts of autonomous logistics sorting systems. Across five open VLMs, \system consistently outperforms matched baselines; for example, the accuracy on SortingBench of Qwen3-VL-8B increases from 63.6\% to 78.8\%. Additional experiments verify the effectiveness of each component and JMSU's spillover benefits in embodied tasks. Deployment tests involving more than 15,000 packages support the practical viability of VLM-based planning for autonomous logistics sorting.
	\end{abstract}

	\section{Introduction}
\label{sec:1_introduction}

\begin{figure}[tbp]
	\centering
	\includegraphics[width=1\columnwidth]{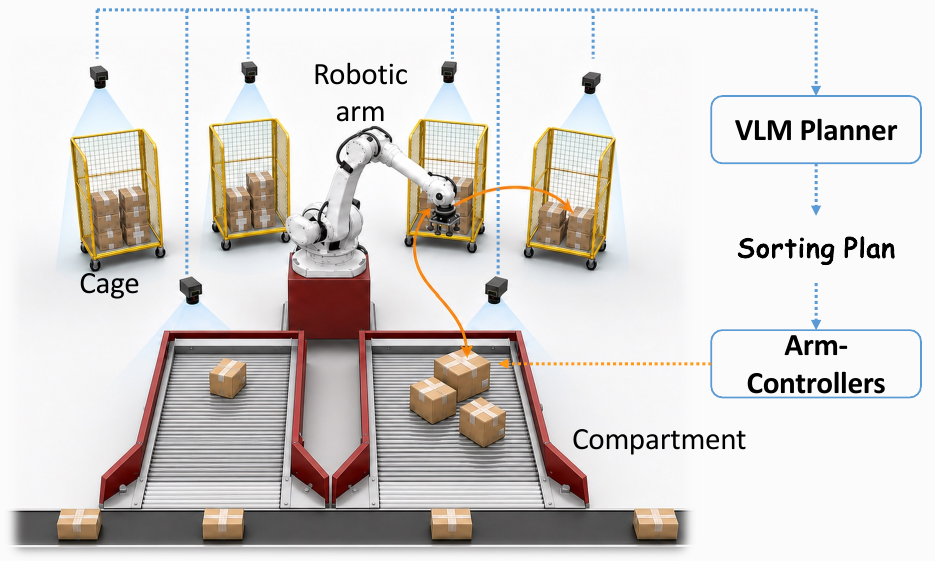}
	\caption{{Illustration of ALSS. Based on the VLM, ALSS enables automated grasping of non-standard packages from compartments, handling, and stacking them into cages, according to diverse business requirements.}}
	\label{fig:scene_and_demo}
\end{figure}

With the advancement of AI, modern logistics workstations increasingly integrate robotic arms, destination cages, distributed cameras, and low-level controllers into an autonomous logistics sorting system (ALSS), as illustrated in Figure~\ref{fig:scene_and_demo}. ALSS continuously perceives its physical environment, plans sorting operations, and executes corresponding actions.
In addition to its direct industrial value, it represents a practical way to bring embodied AI into real industrial operations.   

To perform reliable sorting planning, ALSS requires visual evidence from multiple spatially distributed regions.
A global camera provides broad coverage but loses the fine-grained details required for package localization, whereas a local egocentric camera preserves detail but cannot observe all decision-relevant regions.
ALSS therefore uses multiple cameras to capture complementary observations of distinct physical scenes. 
We abstract this requirement as Joint Multi-Scene Understanding (JMSU), which exhibits two defining characteristics:
\squishlist
\item \textbf{Spatial distribution:} The observations cover physically disjoint regions with little field-of-view overlap.
\item {\textbf{Decision-level interdependency:} Different scenes provide complementary evidence for the final decision, and removing a key image may invalidate the plan.}
\squishend
This combination of spatial distribution and decision-level interdependency makes JMSU a planning problem for ALSS rather than an independent per-image perception problem.

Vision-Language Models (VLMs) are promising candidates for JMSU because they combine broad visual-semantic knowledge, instruction-conditioned reasoning, and structured language generation~\cite{p_saycan_ahn2022can,p_robobrain_ji2025robobrain,p_openvla_kim2024openvla}. A VLM can provide a unified interface for package grounding, cross-scene state comparison, and action generation, while specialized controllers retain responsibility for physical execution. However, enabling VLMs to handle JMSU presents two major challenges due to the gap between general pre-training and JMSU’s embodied requirements:
\squishlist
	\item First, there is a substantial domain discrepancy and a scarcity of high-quality JMSU data. VLMs are primarily pre-trained on web-scale image-text data {\cite{p_vita_fu2025vita,p_ovis_lu2025ovis2,p_qwen2.5_bai2025qwen2}}, which differ substantially from the observations and tasks of ALSS. Although workstations continuously produce abundant raw data, most data are noisy, redundant, or lack valid action annotations, making high-quality JMSU supervision scarce and expensive. Moreover, conventional augmentation that independently modifies individual images or text may break the consistency among distributed observations, physical states, and action labels.
	\item Second, JMSU requires holistic reasoning over long visual contexts. According to the physical distribution of the equipment, the VLM must process multiple high-resolution observations covering all candidate compartments, packages, and destination cages. As the number of visual tokens increases, attention is distributed over a larger context, making it more difficult to identify critical evidence and integrate the complete system state. 
	
\squishend

To address these challenges, we propose \system, a comprehensive optimization scheme for VLM-based JMSU reasoning in ALSS. 
\system contains two complementary components. At the data level, we introduce Endogenous Data Augmentation (EDA) to decompose each annotated sample into verifiable atomic facts and recombine them under domain-specific operational constraints. It generates diverse training tasks while preserving consistency among visual observations, physical states, and action labels. At the model level, we design Global Context Ranking (GCR) as an auxiliary training objective that encourages the instruction representation to align more strongly with the complete visual context than with any partial visual context. GCR therefore guides the VLM to integrate all decision-relevant scenes before producing an action plan, without changing the inference architecture or introducing additional inference-time modules.
%We evaluate \system in an operational logistics sorting environment covering approximately $70m^2$.

Furthermore, over three months, we collect 2,000 JMSU training samples and construct SortingBench with 1,000 evaluation samples under four workstation layouts.
The benchmark requires models to jointly determine the source compartment, localize the target package, select the destination cage, and generate the complete manipulation plan. Its test set includes unseen layouts and lighting conditions to evaluate both sample-level and scene-level generalization. 

Our main contributions are as follows:
\squishlist
	\item {We formulate JMSU as an ALSS-oriented task and identify data scarcity and attention distraction as bottlenecks. Then we construct a real-world JMSU dataset and benchmark.}
	\item We propose \system, which combines constraint-preserving Endogenous Data Augmentation with Global Context Ranking to improve data efficiency and complete-context reasoning without adding inference-time components.
	\item {Experiments across five open VLMs show consistent improvements over matched SFT baselines ($+15.2\%$ on SortingBench for Qwen3-VL-8B), stronger embodied multi-image reasoning, and limited change in general VQA.}
\squishend

%Upon acceptance of this paper, all source code, dataset, and SortingBench will be made publicly available.

	\section{Related Work}
\label{sec:2_relatedworks}

\textbf{Vision-language planning for autonomous systems.}
{VLM-based embodied systems commonly use a foundation model either as a high-level planner that dispatches low-level skills or as the initialization of an end-to-end vision-language-action policy~\cite{p_saycan_ahn2022can,p_code_as_policy_liang2023code,p_robobrain_ji2025robobrain,p_openvla_kim2024openvla,p_rationalvla_song2025rationalvla}. These systems have expanded the range of objects and instructions that robots can handle, but most evaluations assume that the decision-relevant state is available from a single egocentric view or from overlapping views of the same manipulation region. Industrial autonomous systems impose a different sensing constraint: cameras are assigned to physically separated functional regions, and a task-level planner must combine their states before selecting an action. Recent logistics work confirms the value of embodied models for package handling~\cite{p_logistic_xu2025embodied}. 
	Our study complements this line by isolating and evaluating the distributed-perception problem inside ALSS.}

\textbf{Data augmentation for visual-language reasoning.}
{Domain adaptation is necessary when VLM pre-training does not cover industrial objects and operating rules. Existing multimodal augmentation often generates new rationales, rewrites instructions, or adds more supervision~\cite{p_scienceQA_zhang2023multimodal,p_vision_r1_huang2025vision,p_vla-r1_ye2025vla,p_veclip_lai2024veclip,p_allseeing_wang2023all}. These strategies are valuable for individual images, but independently modifying a view or label can invalidate the cross-scene constraints of JMSU. EDA instead recombines verified facts under explicit picking and placement rules, effectively generating logically correct samples. }

\textbf{Multi-image understanding and reasoning.}
{Recent VLMs improve multi-image processing through higher-resolution visual encoding, interleaved position representations, and targeted instruction tuning~\cite{p_qwen3,p_pixtral_agrawal2024pixtral,p_deepseek,p_internvl3.5,p_molmo,p_minicpm_yu2025minicpm}. Other methods explicitly improve image-text association or reduce order-dependent bias~\cite{p_mia-dpo_liu2024mia,p_vega_zhou2024vega,p_sofa_tian2025identifying}. 
	However, these objectives do not explicitly require the model to form a query-relevant representation that jointly integrates all visual inputs. A model may therefore rely on dominant evidence from an individual image without capturing the cross-image context needed for a unified decision. This motivates GCR, which enforces stronger semantic alignment between the instruction query and the global visual context than between the query and local visual features.}
	\section{Problem Setting: JMSU for ALSS}
\label{sec:3_problem}

\subsection{Autonomous Logistics Sorting Loop}
\label{sec:3.1_sorting_scenario}

%\begin{figure*}[tbp]
%    \centering
%    \includegraphics[width=1.83\columnwidth]{figures/scenes_and_demo.pdf}
%    \caption{{The ALSS testbed and its JMSU decision task. (a) Top-down workstation layouts; blue-underlined indices denote sampled regions without destination cages. (b) Physical equipment and distributed camera views. © A VLM receives the synchronized views and produces a structured sorting plan.}}
%    \label{fig:scene_and_demo}
%\end{figure*}

We study an operational ALSS spanning approximately $70 \mathrm{m}^2$. The system contains source compartments, destination cages, distributed RGB-D cameras, a robotic arm, and calibrated low-level controllers, as shown in Figure~\ref{fig:scene_and_demo}.
The VLM operates at the task-planning layer. At each sorting cycle, synchronized observations are passed to the VLM, which selects a source compartment, localizes the package to be grasped, chooses a destination cage, and emits a structured action sequence. The selected actions invoke calibrated perception, motion-planning, grasping, and placement controllers. This separation gives the learned planner access to open-world semantics while retaining deterministic safety checks and stable execution at the control layer.
%Two operating policies couple the distributed observations.
This plan is also constrained by strict business requirements.
First, the system selects the compartment with the most packages and applies first-in-first-out picking by choosing the package closest to that compartment’s exit. Second, it balances load by selecting the cage with the largest remaining capacity.
%A valid plan consequently depends on evidence from both source and destination regions. Per-image recognition alone is insufficient because the plan is defined by comparisons over the complete system state.

\subsection{JMSU as Complete-State Reasoning}
\label{sec:3.2_JMSU_task_definition}
JMSU formalizes the subset of multi-image embodied planning. Mathematically, let $\mathcal{I}=\{I_1,\ldots,I_N\}$ denote synchronized observations, $\mathcal{T}$ the task instruction and operating policy, and $\mathcal{Y}$ the correct task-level decision. A VLM $\mathcal{M}$ predicts $\hat{\mathcal{Y}}=\mathcal{M}(\mathcal{I},\mathcal{T})$. Within $\mathcal{I}$, $\mathcal{I}^{*}$ is a minimal sufficient set of decision-relevant images; the remaining images $\mathcal{I}_{\mathrm{noise}}=\mathcal{I}\setminus\mathcal{I}^{*}$ are distractors. JMSU satisfies:

%\begin{align*}
%    &{\mathcal{I}^{*}\subseteq\mathcal{I},\qquad |\mathcal{I}^{*}|\geq 2,}\tag{1}\label{eq:jmsu-core}\[-0.4ex]
%    &{\mu!\left(\mathcal{P}(I_i),\mathcal{P}(I_j)\right)\leq\epsilon,\quad 0\leq\epsilon\ll1,}\tag{2}\label{eq:jmsu-space}\[-0.4ex]
%    &{\hspace{8em}\forall I_i,I_j\in\mathcal{I},\ i\neq j,}\[-0.4ex]
%    &{H(\mathcal{Y}\mid\mathcal{I}^{*},\mathcal{T})\leq\delta,\qquad \delta\rightarrow0^{+},}\tag{3}\label{eq:jmsu-sufficient}\[-0.4ex]
%    &{H(\mathcal{Y}\mid\mathcal{S},\mathcal{T})\geq\gamma,\quad
	%      \forall\mathcal{S}\subsetneq\mathcal{I}^{*},\quad \gamma>\delta,}\tag{4}\label{eq:jmsu-necessary}
%\end{align*}
\begin{align*}
	&\mathcal{I}^{*}\subseteq\mathcal{I},\quad |\mathcal{I}^{*}|\geq 2,\tag{1}\label{eq:jmsu-core}\\
	&\mu\!\left(\mathcal{P}(I_i),\mathcal{P}(I_j)\right)\leq\epsilon,\ 0\leq\epsilon\ll1,\forall I_i,I_j\in\mathcal{I},\ i\neq j,\tag{2}\label{eq:jmsu-space}\\
	&H(\mathcal{Y}\mid\mathcal{I}^{*},\mathcal{T})\leq\delta,\quad \delta\rightarrow0^{+},\tag{3}\label{eq:jmsu-sufficient}\\
	&H(\mathcal{Y}\mid\mathcal{S},\mathcal{T})\geq\gamma,\quad
	\forall\mathcal{S}\subsetneq\mathcal{I}^{*},\quad \gamma>\delta,\tag{4}\label{eq:jmsu-necessary}
\end{align*}
{where $\mathcal{P}(I_i)$ maps an observation to its physical field of view and $\mu(\cdot,\cdot)$ is intersection-over-union (IoU) measure. Equation~\eqref{eq:jmsu-space} encodes spatial disjointness. Equations~\eqref{eq:jmsu-sufficient}–\eqref{eq:jmsu-necessary} encode decision-level interdependency: the core set nearly determines the decision, whereas every proper subset retains substantial uncertainty~\cite{p_h_achille2018emergence}.
	% This definition identifies a measurable autonomous-planning regime that is obscured when benchmarks contain separable questions, redundant views, or labels recoverable from one image.
}

\subsection{Sorting Plan Interface}
\label{sec:3.3_interface}

\begin{figure}[tbp]
	\centering
	\includegraphics[width=0.7\columnwidth]{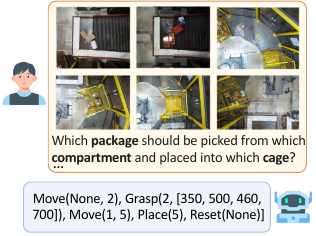}
	\caption{A JMSU sample.  The VLM receives the synchronized views and produces a structured sorting plan.}
	\label{fig:jmsu_qa}
\end{figure}

For ALSS, each image depicts a compartment, a destination cage, or an in-domain distractor.
A complete JMSU sample is shown in Figure~\ref{fig:jmsu_qa}.
The task output is an ordered sequence over $\mathcal{A}=\{\mathrm{Move},\mathrm{Grasp},\mathrm{Place},\mathrm{Reset}\}$. $\mathrm{Move}(a,b)$ transfers the arm from region $a$ to region $b$; $\mathrm{Grasp}(a,B)$ selects the package localized by bounding box $B$ in region $a$; $\mathrm{Place}(a)$ deposits the package into cage $a$; and $\mathrm{Reset}(\varnothing)$ returns the system to its initial state.
A prediction is successful only if the entire action sequence is correct and the package bounding box reaches the required overlap threshold.

	\section{Method}
\label{sec:4_methods}
\begin{figure*}[tbp]
	\centering
	\includegraphics[width=2\columnwidth]{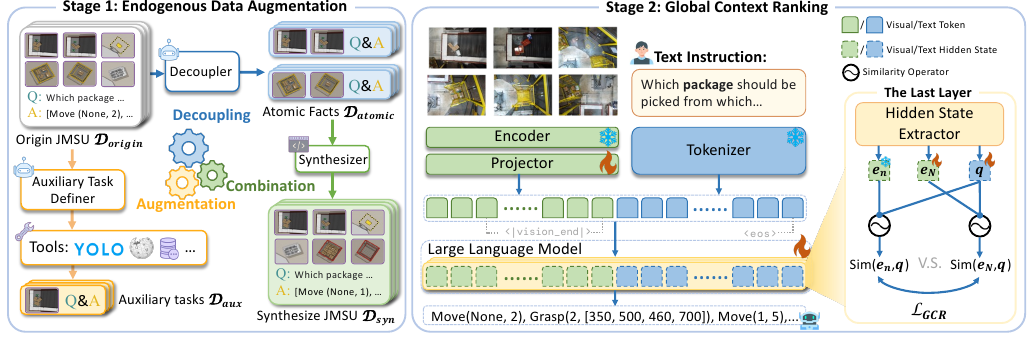}
	% \caption{{Overview of \system. EDA expands limited ALSS supervision by recombining atomic facts under operating constraints. GCR then trains the VLM to align the instruction with the complete visual context more strongly than with the partial.}}
	\caption{A unified framework to enhance VLMs on JMSU. Hugin includes two stages: EDA to mitigate the scarcity of JMSU data, and GCR to achieve semantic alignment between global visual context and instruction query.}
	\label{fig:overview}
\end{figure*}

{We introduce \system to adapt VLMs to JMSU. Figure~\ref{fig:overview} summarizes its two complementary training components. EDA converts limited ALSS annotations into constraint-consistent supervision, addressing data scarcity. GCR enforces a stronger semantic alignment between the instruction query and the global visual context than between query and local features, preventing the VLM from relying on local visual prefixes.}

\subsection{Endogenous Data Augmentation}
\label{sec:4.1_EDA}

As noted in Section~\ref{sec:1_introduction}, fine-tuning VLMs is severely bottlenecked by the scarcity and prohibitive collection costs of JMSU data.
While data augmentation offers a natural solution, standard techniques typically disrupt the logical dependencies across scenes. For instance, CoT-based methods~\cite{p_scienceQA_zhang2023multimodal,p_vision_r1_huang2025vision,p_vla-r1_ye2025vla} rely on teacher models for synthesis, which risks introducing hallucinations without domain priors. Text-centric methods~\cite{p_llava,p_veclip_lai2024veclip} fail to scale the visual information density~\cite{p_svit_zhao2023svit}.
Instead of relying on external generation, \system introduces a data augmentation method named EDA, as shown in Stage $1$ of Figure~\ref{fig:overview}.
EDA exploits the inherent semantic richness of existing data~\cite{p_rephrase_deng2023rephrase}.
By treating complex tasks as recombinable atomic facts rather than indivisible units~\cite{p_gqa_hudson2019gqa}, it polynomially expands the effective training distribution while maintaining logical consistency.
Specifically, EDA consists of the following three stages.

%\begin{algorithm}[bp]
%    \caption{decoupling and combination for JMSU tasks}
%    \label{alg:synthetic_jmsu}
%    \KwIn{JMSU Dataset $\mathcal{D}_{origin}$, LLM Parser $\mathcal{M}$, Target size $K$.}
%    \KwOut{Atomic Facts $\mathcal{D}_{atomic}$, Synthetic Dataset $\mathcal{D}_{syn}$.}
%    % \tcc{Phase 1: Decomposition}
%    $s{ref} \sim \textsc{Sample}(\mathcal{D}_{origin})$ \tcp*{Select a reference sample}
%    ${T_i}_{i=1}^N \gets \textsc{ParseFact}(\mathcal{M}, s_{ref})$ \tcp*{One-time fact induction}
%    Initialize fact pools ${F_i}{i=1}^N \gets \emptyset$, $\mathcal{D}{syn} \gets \emptyset$;
%    \ForEach{sample $s \in \mathcal{D}{origin}$}{
	%        ${f_i}{i=1}^N \gets \textsc{Decoupler}(\mathcal{M}, s, {T_i}{i=1}^N)$;
	%        \For{$i = 1$ \KwTo $N$}{
		%            $F_i \leftarrow F_i \cup {f_i}$ \tcp*{Add $f_i$ to corresponding fact pool}
		%        }
	%    }
%    % \tcc{Phase 2: Synthesis}
%    \For{$k = 1$ \KwTo $K$}{
	%        $\text{facts} \leftarrow \emptyset$;
	%        \For{$i = 1$ \KwTo $N$}{
		%            $\text{facts} \leftarrow \text{facts} \cup {\textsc{Sample}(F_i)}$;
		%        }
	%        $S \gets \textsc{Synthesizer}(\text{facts})$ \tcp*{Synthesize a JMSU sample}
	%        $\mathcal{D}{syn} \leftarrow \mathcal{D}{syn} \cup {S}$;
	%        % \If{$\textsc{Verify}(S)$}{
		%        %     $\mathcal{D}{syn} \leftarrow \mathcal{D}{syn} \cup {S}$;
		%        % }
	%    }
%    \Return ${F_i}{i=1}^N$, {$\mathcal{D}_{syn}$};
%\end{algorithm}

\textbf{{Decoupling: semantic decomposition into atomic facts}.}
{Given the original dataset $\mathcal{D}_{origin}$, the LLM-based decoupler parses each sample into the shared fact schema and places the extracted facts in $\mathcal{D}_{atomic}$. In ALSS, the fact types include package counts, the package closest to a compartment exit, cage occupancy, and spatial grounding. The associated atomic questions expose intermediate supervision that is hidden inside the final action sequence.}

\textbf{{Combination: constraint-based task synthesis}.}
{The script-based synthesizer samples compatible facts from the pools and recomputes a complete plan under the operating constraints, producing $\mathcal{D}_{syn}$~\cite{p_ProcTHOR_deitke2022}. If a sample contains $M$ fact types and each pool contains at most $N$ candidates, the candidate space is bounded by $N^{M}$; EDA materializes only a balanced subset for training. Because the answer is computed from structured facts rather than copied from a teacher VLM, the label remains auditable. We additionally inspect randomly sampled synthetic records and verify their structure and labels before training.}

% \begin{algorithm}[htbp]
	%     \caption{Synthesize a JMSU sample from atomic fact classes}
	%     \label{alg:synthetic_jmsu}
	%     \begin{algorithmic}[1]
		%     \Require Atomic fact classes ${F_1,$ $ F_2, $ $\dots,$ $F_N}$.
		%     \Ensure A synthetic JMSU sample $S$.
		%     \State Initialize an empty list $\text{selected_facts} \gets [,]$
		%     \For{$k = 1$ to $N$}
		%         \State Sample $f_{k,i_k} \sim \text{Uniform}(F_k)$  \Comment{Randomly pick one fact}
		%         \State Append $f_{k,i_k}$ to $\text{selected_facts}$
		%     \EndFor
		%     \State $S \gets \textsc{Combine}(\text{selected_facts})$  \Comment{Fuse facts into a JMSU sample}
		%     \State \Return $S$
		%     \end{algorithmic}
	% \end{algorithm}

\textbf{{Augmentation: perception and general-VQA supervision}.}
EDA also derives $\mathcal{D}_{aux}$ for object detection, region identification, and cage-state recognition using specialized vision tools~\cite{p_allseeing_wang2023all,p_allseeing2_wang2024all,p_yolo_hussain2023yolo,p_sam_kirillov2023segment}. These tasks ground the planner’s comparative decisions in package and cage perception. General VQA samples $\mathcal{D}_{general}$ are mixed into training to reduce catastrophic forgetting~\cite{p_coco_liu2024improved}.
The final training set is $\mathcal{D}=\mathcal{D}_{origin}\cup\mathcal{D}_{atomic}\cup\mathcal{D}_{syn}\cup\mathcal{D}_{aux}\cup\mathcal{D}_{general}$, which helps VLMs to learn robust and grounded representations from limited samples.

The complete EDA implementation and processing results are provided in the supplementary material.

\begin{figure}[tbp]
	\centering
	\includegraphics[width=1\columnwidth]{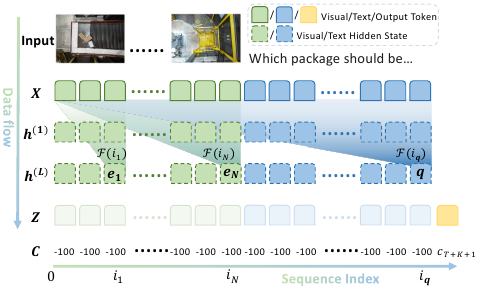}
	\caption{The data flow process during VLM training.}
	\label{fig:dataflow}
\end{figure}

\subsection{Global Context Ranking}
\label{sec:4.2_GCR}
A decoder-only VLM serializes simultaneous camera observations, so standard next-token training may produce the correct label without explicitly preferring a representation of the complete visual state.
GCR introduces that preference as a training-only ranking objective. Based on the information aggregation property of causal attention~\cite{p_transformer_vaswani2017attention}, GCR utilizes specific hidden states as anchors for JMSU task understanding. To further improve global awareness, GCR enforces a stronger semantic alignment the instruction query more strongly with the global visual context than with local visual features, similar to contrastive learning~\cite{p_simcse_gao2021simcse, p_coin_chen2024coin}.

\textbf{Theoretical basis}. GCR intrinsically exploits the unidirectional information flow governed by the causal attention mechanism in decoder-only VLMs and the inherent information aggregation property of the transformer. Given the multimodal input $[I_1,$ $I_2,$ $\dots,$ $I_N,$ $\mathcal{T}]$, let it be tokenized and flattened into a prefill token sequence $\boldsymbol{X} = [x_1, x_2, \dots, x_T]$ of length $T$, where $x_t$ denotes the $t$-th token. In the $l$-th transformer layer, the multi-head attention updates $x_t$’s hidden state $\boldsymbol{h}_t^{(l)} \in \mathbb{R}^d$ via a lower-triangular causal mask $\boldsymbol{M} \in \mathbb{R}^{T \times T}$, where $\boldsymbol{M}_{t,\tau} = 0$ if $\tau \leq t$ and $- \infty$ otherwise. This physically blocks any backward propagation of information, ensuring the contextual receptive field $\mathcal{F}(t)$ of $\boldsymbol{h}_t^{(l)}$ is rigorously bounded to its historical token prefix, \ie $\mathcal{F}(t) = \{x_1, \dots, x_t\}$, as shown in Figure~\ref{fig:dataflow}.
Furthermore, this unidirectional masking induces information aggregation~\cite{p_transformer_vaswani2017attention}.
The hidden state $\boldsymbol{h}_t^{(l)}$ acts as a summary, mathematically aggregating the entire multimodal history $x_{1:t}$ into a single dense vector~\cite{p_aligning_kong2024aligning,p_hiddenstate_valeriani2023geometry,p_hiprobe_cai2025hiprobe}. Therefore, we can treat the hidden states of specific tokens as precise semantic anchors which capture the VLM’s cognitive states at different sequence positions during the prefill stage~\cite{p_simcse_gao2021simcse,p_context_hendel2023context}.

\textbf{Anchor indexing.}
To operationalize the extraction of specific cognitive states, we need to locate these valuable anchors. Let $\mathcal{E}_{vis}$ denote the ordered set of positional indices corresponding to the boundary identifiers of images (\eg \texttt{<|vision\_end|>} for Qwen3-VL~\cite{p_qwen3}:  % \texttt{<\textbackslash image>} for MiniCPM-V4_5~\cite{p_minicpm_yu2025minicpm}
\begin{equation*}
\begin{aligned}
	\mathcal{E}_{vis} &= \{ t \in \{1, \dots, T\} \mid x_t = \texttt{<|vision\_end|>} \} \\
	&= \{i_1 < \dots < i_N\}.
\end{aligned}
\end{equation*}
%\begin{equation*}
% \mathcal{E}_{\mathrm{vis}}
% ={t\in{1,\ldots,T}\mid x_t=\texttt{<|vision_end|>}}
% ={i_1<\cdots<i_N}.
%\end{equation*}
Similarly, during the decoding stage, the VLM expands $\boldsymbol{X}$ to sequence $\boldsymbol{Z}$. Since different VLMs employ varying chat templates that append specific generation triggers (\eg \texttt{<|im\_start|>}~\cite{p_qwen3}), we denote these tokens as $\boldsymbol{S}$ $=$ $[s_1,$ $\dots,$ $s_K]$. Therefore, $\boldsymbol{Z}$ is formulated as $\boldsymbol{Z} = [\boldsymbol{X}, \boldsymbol{S}, \boldsymbol{Y}]$ (ignoring prefix system prompts), where $\boldsymbol{Y} = [y_1, \dots, y_{T'}]$ is the sequence of generated answer tokens.
Let $\boldsymbol{C} = [c_1, \dots, c_{T+K+T'}]$ be the corresponding label sequence, where the tokens of $\boldsymbol{X}$ and $\boldsymbol{S}$ are masked by an ignore-index (\eg $-100$).
To adapt to the differences among chat templates, we anchor the instruction intent $\boldsymbol{q}$ at the token immediately preceding trainable answer tokens, and robustly define its index $i_{\boldsymbol{q}}$ as follows:
% Some VLM architectures may add additional flags or system prompts at the end of $X$.
% For instance,xxxxxxxxxxxxxx %【添加引用和举例，运行swift看看】
% Therefore, to ensure generality, we define $q$’s index $p_q$ as preceding the first answer token:
\begin{equation*}
i_{\boldsymbol{q}}=\min\{t\mid 1\leq t\leq T+K+T',
	c_t\in\boldsymbol{C}\setminus\{-100\}\}-1.
	\end{equation*}
	
	\textbf{Hidden state extraction.}
	Using these indices, we extract 3 critical representations from the final hidden state $\boldsymbol{H}^{(L)}$:
	\squishlist
	\item \textbf{Local visual context} $\boldsymbol{e}_n \in \mathbb{R}^d$: Extracted by uniformly sampling an intermediate visual boundary, $\boldsymbol{e}_n = \boldsymbol{h}_{i_n}^{(L)}$, where $n \sim \mathcal{U}\{1, N-1\}$. Governed by the causal mask, its receptive field is structurally restricted to $\mathcal{F}(i_n) = \{x_1, x_2, \dots, x_{i_n}\}$. This hidden state encapsulates the tokenized representations of the first $n$ images, and is blind to the subsequent visual tokens of $I_{n+1:N}$.
	\item \textbf{Global visual context} $\boldsymbol{e}_N \in \mathbb{R}^d$: Extracted at the final visual boundary, $\boldsymbol{e}_N = \boldsymbol{h}_{i_N}^{(L)}$. Its receptive field $\mathcal{F}(i_N) = \{x_1, x_2, \dots, x_{i_N}\}$ encompasses the tokenized visual history of all $N$ images.
	\item \textbf{Instruction intent} $\boldsymbol{q} \in \mathbb{R}^d$: Extracted at the prefill boundary, $\boldsymbol{q} = \boldsymbol{h}_{i_{\boldsymbol{q}}}^{(L)}$. Its receptive field $\mathcal{F}(i_{\boldsymbol{q}})$ aggregates all visual and textual context. So it represents the model’s final understanding of the task~\cite{p_coin_chen2024coin}.
	\squishend
	
	\textbf{GCR optimization.}
	To translate the structural hierarchy of visual contexts into a learnable objective, we introduce the GCR.
	Given the topological subset relation of their token spans, \ie $\mathcal{F}(i_n) \subsetneq \mathcal{F}(i_N) \subsetneq \mathcal{F}(i_{\boldsymbol{q}})$, if the VLM effectively integrates multi-scene information, the semantic alignment of $\boldsymbol{q}$ and $\boldsymbol{e}_N$ should exceed that between $\boldsymbol{q}$ and $\boldsymbol{e}_n$.
	Therefore, we project these hidden states onto a unit hypersphere, \ie $\tilde{\boldsymbol{v}} = \boldsymbol{v} / ||\boldsymbol{v}||_2$ for $\boldsymbol{v} \in \{\boldsymbol{q}, \boldsymbol{e}_N, \boldsymbol{e}_n\}$, adopt cosine similarity to evaluate the alignment, and optimize the expected margin-based Hinge loss~\cite{p_hingeloss_gentile1998linear}:
	\begin{equation*}
{\begin{aligned}
		\mathcal{L}_{GCR}=\mathbb{E}_{n\sim\mathcal{U}\{1,N-1\}}\big[\max(0,\alpha
		+\mathcal{SG}(\tilde{\boldsymbol{e}}_n)^\top\tilde{\boldsymbol{q}} -\tilde{\boldsymbol{e}}_N^\top\tilde{\boldsymbol{q}})\big],
\end{aligned}}
\end{equation*}
where $\alpha$ is a margin and $\mathcal{SG}(\cdot)$ denotes the Stop-Gradient operator.
During backpropagation, $\mathcal{SG}(\cdot)$ explicitly detaches $\tilde{\boldsymbol{e}}_n$ from the computational graph to serve as a fixed negative. This encourages the model to enhance $\boldsymbol{e}_N$ rather than degrading $\boldsymbol{e}_n$.
This prevents the model from learning trivial solutions that would damage its core visual perception capabilities.
{We set a small margin $\alpha=0.1$ to favor complete-context alignment without forcing the model to discard local evidence needed for grounding.}
The final training objective combines the standard Cross-Entropy (CE) loss with the dynamically gated GCR constraint, formulated as:
\begin{equation*}
\mathcal{L}_{total} = \mathcal{L}_{CE} + \lambda \cdot \mathbb{I}(N \geq 2) \cdot \mathcal{L}_{GCR},
\end{equation*}
{where $\mathbb{I}(N\geq2)$ activates GCR only for multi-image samples, and $\lambda$ controls its contribution relative to next-token prediction. GCR is removed after training; inference cost and the base VLM’s input–output interface are unchanged.}
	\vfill
\pagebreak
% 新开半栏

\section{Experiments}
\label{sec:4_experiments}

\subsection{Data, Baselines, and Metrics}
\label{sec:4.1_setup}

\textbf{{Real-world data and SortingBench}.}
We collected JMSU samples from 4 ALSS workstations over 3 months. The domain-adaptation set $\mathcal{D}_{origin}$ contains 2,000 samples from Layouts 1–2. Each sample includes one or two in-domain distractor views, and all images are randomly permuted. SortingBench contains 1,000 held-out samples from 4 layouts.

\begin{table*}[!t]
	\centering
	\setlength{\tabcolsep}{0pt}
	\resizebox{\textwidth}{!}{%
		\begin{tabularx}{1.1\textwidth}{@{}l|*{3}{C}|*{4}{C}|*{5}{C}@{}}
			\toprule
			\multirow{2}[4]{*}{\textbf{Model}}
			& \multicolumn{3}{c|}{\textbf{General}}
			& \multicolumn{4}{c|}{\textbf{Embodied-related multi-image}}
			& \multicolumn{5}{c}{\textbf{SortingBench}} \\
			\cmidrule{2-13}
			& {\scriptsize\textbf{MMB/EN}} & {\scriptsize\textbf{MMB/CN}} & {\scriptsize\textbf{MME}}
			& {\textbf{OL}} & {\textbf{VS}} & {\textbf{Co.}} & {\textbf{SU}}
			& {\textbf{L1}} & {\textbf{L2}} & {\textbf{L3}} & {\textbf{L4}} & \textbf{Overall} \\
			\midrule
			\rowcolor{gray!12}\multicolumn{13}{c}{\textbf{Closed-source VLMs}} \\
			\midrule
			{GPT-5.5} & {-} & {-} & {--} & {--} & {--} & {--} & {--} & {0.0} & {2.8} & {0.0} & {0.0} & {0.7} \\
			{GPT-5-nano-high} & {78.9} & {79.6} & {1426} & {41.0} & {80.7} & {60.0} & {78.0} & {0.0} & {0.0} & {0.0} & {0.0} & {0.0} \\
			\midrule
			\rowcolor{gray!12}\multicolumn{13}{c}{\textbf{Open-source VLMs}} \\
			\midrule
			{Ovis2.5-2B} & {\underline{80.5}} & {\underline{76.8}} & {\underline{1585}} & {\underline{58.2}} & {\textbf{81.5}} & {\underline{68.3}} & {\underline{75.3}} & {0.0} & {0.0} & {0.0} & {0.0} & {0.0} \\
			{\quad +SFT/\textcolor{softgreen}{+Ego3D}} & {80.4} & {\underline{76.8}} & {\textbf{1603}} & {57.4} & {71.9} & {\textbf{70.8}} & {\textbf{82.8}} & {\underline{57.3}} & {\underline{63.9}} & {\underline{66.3}} & {\underline{51.4}} & {\underline{59.7}/\textcolor{softgreen}{48.7}} \\
			{\quad +\system} & {\textbf{81.5}} & {\textbf{77.1}} & {1568} & {\textbf{59.0}} & {\underline{74.8}} & {61.7} & {74.2} & {\textbf{68.6}} & {\textbf{68.9}} & {\textbf{73.3}} & {\textbf{64.8}} & {\textbf{68.9}} \\
			\midrule
			{Gemma3-4B} & {\underline{66.6}} & {\underline{66.7}} & {\textbf{1384}} & {\underline{54.9}} & {\underline{64.4}} & {\textbf{35.8}} & {--} & {0.0} & {0.0} & {0.0} & {0.0} & {0.0} \\
			{\quad +SFT/\textcolor{softgreen}{+Ego3D}} & {63.0} & {59.8} & {\underline{1346}} & {48.4} & {31.9} & {21.7} & {--} & {\underline{23.0}} & {\underline{29.5}} & {\underline{17.8}} & {\underline{18.6}} & {\underline{22.1}/\textcolor{softgreen}{11.8}} \\
			{\quad +\system} & {\textbf{68.5}} & {\textbf{67.1}} & {1319} & {\textbf{59.8}} & {\textbf{71.9}} & {\underline{30.8}} & {--} & {\textbf{46.8}} & {\textbf{53.9}} & {\textbf{57.4}} & {\textbf{49.4}} & {\textbf{51.9}} \\
			\midrule
			{MiniCPM-V4\_5} & {\underline{84.3}} & {\textbf{84.4}} & {\underline{1720}} & {\underline{57.4}} & {\underline{92.6}} & {\textbf{70.8}} & {--} & {0.0} & {0.0} & {0.0} & {0.0} & {0.0} \\
			{\quad +SFT/\textcolor{softgreen}{+Ego3D}} & {\textbf{85.3}} & {\textbf{84.4}} & {\textbf{1732}} & {56.6} & {91.1} & {\textbf{70.8}} & {--} & {\underline{50.4}} & {\underline{54.4}} & {\underline{55.4}} & {\underline{50.2}} & {52.6/\textcolor{softgreen}{\underline{52.9}}} \\
			{\quad +\system} & {83.8} & {\underline{82.9}} & {1674} & {\textbf{58.2}} & {\textbf{94.1}} & {\underline{66.7}} & {--} & {\textbf{78.2}} & {\textbf{70.5}} & {\textbf{77.1}} & {\textbf{74.3}} & {\textbf{75.1}} \\
			\midrule
			{Qwen3-VL-4B} & {\textbf{82.9}} & {\textbf{81.9}} & {\textbf{1705}} & {\textbf{68.0}} & {79.3} & {\underline{66.7}} & {\underline{67.7}} & {0.0} & {0.0} & {0.0} & {0.0} & {0.0} \\
			{\quad +SFT/\textcolor{softgreen}{+Ego3D}} & {\underline{81.4}} & {\underline{80.9}} & {\underline{1676}} & {61.5} & {\underline{82.2}} & {65.8} & {66.1} & {\underline{65.3}} & {\underline{78.8}} & {\underline{77.1}} & {\underline{70.8}} & {\underline{73.0}/\textcolor{softgreen}{68.4}} \\
			{\quad +\system} & {79.4} & {78.9} & {1633} & {\underline{66.4}} & {\textbf{86.7}} & {\textbf{67.5}} & {\textbf{70.5}} & {\textbf{74.6}} & {\textbf{83.0}} & {\textbf{83.3}} & {\textbf{71.5}} & {\textbf{78.1}} \\
			\midrule
			{RoboBrain2.5-8B-NV} & {77.2} & {80.8} & {1640} & {64.8} & {83} & {62.5} & {67.7} & {0.0} & {0.0} & {0.0} & {0.0} & {\underline{0.0}} \\
			{\quad +SFT} & {67.3} & {73.5} & {1674} & {56.6} & {83.7} & {67.5} & {65.6} & {72.2} & {81.3} & {79.1} & {71.9} & {76.1} \\
			\midrule
			{Qwen3-VL-8B} & {\textbf{83.8}} & {\textbf{83.6}} & {\underline{1734}} & {68.0} & {\underline{85.2}} & {\textbf{68.3}} & {60.8} & {0.0} & {0.0} & {0.0} & {0.0} & {0.0} \\
			{\quad +SFT/\textcolor{softgreen}{+Ego3D}} & {\underline{82.8}} & {81.3} & {1699} & {\textbf{71.3}} & {75.6} & {66.7} & {\underline{64.0}} & {\underline{58.1}} & {\underline{70.5}} & {\underline{68.2}} & {\underline{57.7}} & {\underline{63.6}/\textcolor{softgreen}{38.8}} \\
			{\quad +\system} & {81.4} & {\underline{81.5}} & {\textbf{1743}} & {\underline{69.7}} & {\textbf{88.2}} & {\underline{67.5}} & {\textbf{64.5}} & {\textbf{75.8}} & {\textbf{80.1}} & {\textbf{85.9}} & {\textbf{74.3}} & {\textbf{78.8}} \\
			\bottomrule
	\end{tabularx}}
	\caption{Benchmark Results. Bold indicates top performance; underline, second-best. ``-'' denotes unsupported by VLMEvalKit.}
	\label{tab:performance}
	
	\vspace{6pt}
	\setlength{\tabcolsep}{6pt}
	\begin{tabular}{l|c|cc|cc|c}
		\toprule
		{\textbf{Training}} & \textbf{Original} & \textbf{Shuffle-1} & \textbf{Shuffle-2} & \textbf{+6 distractors} & \textbf{+12 distractors} & \textbf{One-image removal} \\
		\midrule
		{SFT} & {63.6} & {63.3} & {63.8} & {45.8} & {18.8} & {35.6} \\
		{$\mathcal{D}_{origin}$ + GCR} & {68.8} & {69.2} & {68.4} & {55.9} & {34.0} & {--} \\
		{\system} & {\textbf{78.8}} & {\textbf{78.8}} & {\textbf{78.9}} & {\textbf{76.8}} & {\textbf{69.8}} & {\textbf{41.6}} \\
		\bottomrule
	\end{tabular}
	\caption{{SortingBench stress tests on Qwen3-VL-8B. Shuffle-1/2 use two independent permutations. The removal test drops one randomly selected key image; each label depends on two of four or five key images.}}
	\label{tab:stress}
\end{table*}

\textbf{{Training data and implementation}.}
EDA parses the 2,000 JMSU samples into 6,000 atomic facts.
%and can enumerate a pool of 400,000 constraint-consistent JMSU candidates.
{\system}’s training uses approximately 20,000 additional samples: 13,954 atomic or auxiliary perception tasks, 2,000 synthetic JMSU tasks generated by EDA, and 4,000 general VQA samples from LLaVA-1.5~\cite{p_coco_liu2024improved}. We fine-tune the projector and language model for $2$ epochs on $4$ NVIDIA L20 GPUs with learning rate $10^{-5}$. GCR uses $\alpha=0.1$ and $\lambda=0.01$. Base, SFT, and \system variants share the same architecture and optimization schedule, and SFT uses $\mathcal{D}_{origin}$ only.

\textbf{{Baselines and metrics}.}
We evaluate five open VLMs (Ovis2.5-2B, Gemma3-4B, MiniCPM-V4\_5, and Qwen3-VL-4B/8B) and two GPT models (GPT-5.5, GPT-5-nano-high). We also compare with RoboBrain2.5-8B-NV, a recent embodied VLM trained for spatial and temporal reasoning~\cite{p_robobrain25_tan2026}, and with Ego3D-VLM~\cite{p_ego3d_gholami2025spatialreasoningvisionlanguagemodels}, which generates cognitive maps to support egocentric 3D reasoning.
%A SortingBench prediction is correct only if its full action sequence is correct and the predicted package box has IoU above $85%$.
Furthermore, we select 4 embodied tasks from established multi-image benchmarks to assess how capabilities on JMSU affect VLMs’ transferability to broader embodied tasks, including Object-Localization (OL), Visual-Similarity (VS), BLINK Counting (Co.) from BLINK ~\cite{p_blink_fu2024blink}, and Scene-Understanding (SU) from MUIRBench~\cite{p_muirbench_wang2024muirbench}. To monitor the impact on general capabilities, we also evaluate VLMs on general VQA benchmarks, including MMBench\_TEST\_EN\_V11 (MMB/EN), MMBench\_TEST\_CN\_V11 (MMB/CN)~\cite{p_mmbench_liu2024mmbench} and MME~\cite{p_mme_fu2025mme}.
All public benchmarks use VLMEvalKit~\cite{p_vlmevalkit_duan2024vlmevalkit}.
The IoU threshold of bounding boxes for detecting packages is $85\%$.

More details are provided in the supplementary material.

\subsection{Main Results}
\label{sec:4.2_benchmark_results}

\textbf{{Necessary-image intervention}.} We first examine the JMSU features of SortingBench. {If the removed image is sampled uniformly from four or five key views and the decision depends on two of them, the expected retained success is $[1/2,3/5]$ of the original rate. Experimentally, SFT falls from $63.6\%$ to $35.6\%$ (expected interval $[31.8,38.2]\%$), and \system falls from $78.8\%$ to $41.6\%$ (expected interval $[39.4,47.3]\%$). The intervention behaves as predicted by the minimal-sufficient-set definition of Equation~\ref{eq:jmsu-necessary}.}

\textbf{{Domain knowledge alone does not solve complete-state reasoning}.}
{Table~\ref{tab:performance} shows that general VLMs score near zero on SortingBench despite strong general-VQA results. SFT establishes a domain-aware planner, but \system improves all five open VLMs over their matched SFT baselines by $+9.2\%$, $+29.8\%$, $+22.5\%$, $+5.1\%$, and $+15.2\%$, respectively. Ego3D-VLM does not consistently improve the matched SFT models (e.g., $38.8\%$ versus $63.6\%$ on SFT Qwen3-VL-8B). Despite being trained on 12.4M embodied samples, RoboBrain2.5 also scores zero without adaptation and reaches $76.1\%$ after the same 2,000-sample SFT, still below {\system}’s $78.8\%$ with only 22k samples.}

\textbf{{The improvement is consistent across seen and held-out layouts}.}
For Qwen3-VL-8B, \system achieved gains of $17.7\%$, $9.6\%$, $17.7\%$, and $16.6\%$, respectively, across the 4 layouts.
%\system reaches $75.8%$, $80.1%$, $85.9%$, and $74.3%$ on Layouts 1–4, compared with $58.1%$, $70.5%$, $68.2%$, and $57.7%$ for matched SFT. The corresponding gains are $17.7%$, $9.6%$, $17.7%$, and $16.6%$.
Layouts 3–4 are absent from the training set, so their improvements provide direct evidence of held-out-layout transfer within the same logistics application.

\textbf{{The gain transfers beyond sorting without a uniform loss of general ability}.}
On Qwen3-VL-8B, \system improves BLINK/VS from $75.6\%$ to $88.2\%$ and MUIRBench/SU from $64.0\%$ to $64.5\%$. These results show that spatial reasoning for JMSU can yield a positive spillover effect on basic embodied tasks.
Meanwhile, on Qwen3-VL-8B, \system improves MME from $1699$ to $1743$, and MMBench/CN changes from $81.3\%$ to $81.5\%$. These results
demonstrate that general abilities and embodied capabilities in VLMs can be decoupled and co-evolve.
By combining endogenous data augmentation with proper structural constraints, the model can achieve substantial improvements in specific embodied tasks while preserving its general multimodal performance.

%\textbf{{Cumulative success rate identifies the remaining bottleneck}.}
%{A step-wise decomposition further localizes the source of failure. For Qwen3-VL-8B, cumulative success is $95.4%$ after source-compartment selection and $92.9%$ after package grounding, but falls to $78.8%$ when the destination cage must be selected; the corresponding SFT values are $93.9%$, $87.0%$, and $63.6%$. The destination-comparison step therefore causes a $23.4$-point drop for SFT but only a $14.1$-point drop for \system. This step requires comparing spatially disjoint cage states after the source package has already been grounded, confirming global state comparison as the dominant residual error and motivating the controlled tests in the next subsection.}

\subsection{Diagnostic Tests and Ablations}
\label{sec:4.3_ablation}

%\begin{table}[tbp]
%  \centering
%  \caption{{GCR on $\mathcal{D}_{origin}$ only.}}
%  \label{tab:gcr_only}
%  %    \setlength{\tabcolsep}{5pt}
%  \begin{tabular}{lrr}
%  \toprule
%  {Model} & {CE} & {CE+GCR} \
%  \midrule
%  {Qwen3-VL-8B} & {63.6} & \textbf{68.8} \
%  {MiniCPM-V4_5} & {59.7} & \textbf{65.7} \
%  {Ovis2.5-2B} & {52.6} & \textbf{58.7} \
%   \bottomrule
%  \end{tabular}
%\end{table}
%
%\begin{table}[tbp]
%  \centering
%  \caption{{Synthetic JMSU scaling on Qwen3-VL-8B.}}
%  \label{tab:syn_scale}
%  \setlength{\tabcolsep}{4pt}
%  \begin{tabular}{lrrrr}
%  \toprule
%  {$|\mathcal{D}_{syn}|$} & {0} & {2k} & {8k} & {18k} \
%  \midrule
%  {SortingBench} & {63.6} & {75.2} & {82.3} & {83.9} \
%  {Time (min)} & {55} & {102} & {255} & {533} \
%   \bottomrule
%  \end{tabular}
%\end{table}

\begin{table*}[tbp]
\centering
\setlength{\tabcolsep}{1pt}
\begin{tabularx}{\textwidth}{*{4}{C}|*{2}{C}|*{3}{C}|C}
	\toprule
		\multirow{2}[4]{*}{$\mathbf{\mathcal{D}_{emb}}$}
	& \multirow{2}[4]{*}{$\mathbf{\mathcal{D}_{gen}}$}
		& \multirow{2}[4]{*}{\textbf{CE}}
		& \multirow{2}[4]{*}{\textbf{GCR}}
	& \multicolumn{2}{c|}{\textbf{General}}
	& \multicolumn{3}{c|}{\textbf{Embodied-related multi-image}}
		& \multirow{2}[4]{*}{\scriptsize{\textbf{SortingBench}}} \\
	\cmidrule{5-9}
	& & & & \small{\textbf{MMB/EN}} & \small{\textbf{MME}}
		& \small{\textbf{OL}} & \small{\textbf{VS}} & \small{\textbf{Co.}} & \\
	\midrule
		\textcolor{softred}{\ding{55}} & \textcolor{softred}{\ding{55}} & \textcolor{softgreen}{\ding{51}} & \textcolor{softgreen}{\ding{51}} & \textbf{83.0} & \underline{1736} & \textbf{72.1} & 73.3 & \underline{68.3} & 68.8 \\
		\textcolor{softred}{\ding{55}} & \textcolor{softgreen}{\ding{51}} & \textcolor{softgreen}{\ding{51}} & \textcolor{softgreen}{\ding{51}} & 81.0 & 1701 & 63.1 & 83.7 & \textbf{70.0} & 73.3 \\
		\textcolor{softgreen}{\ding{51}} & \textcolor{softred}{\ding{55}} & \textcolor{softgreen}{\ding{51}} & \textcolor{softgreen}{\ding{51}} & 77.3 & 1661 & \underline{69.7} & \underline{84.4} & 57.5 & \underline{78.4} \\
		\textcolor{softgreen}{\ding{51}} & \textcolor{softgreen}{\ding{51}} & \textcolor{softgreen}{\ding{51}} & \textcolor{softred}{\ding{55}} & 80.7 & 1631 & 68.9 & 84.3 & \underline{68.3} & 77.2 \\
		\textcolor{softgreen}{\ding{51}} & \textcolor{softgreen}{\ding{51}} & \textcolor{softgreen}{\ding{51}} & \textcolor{softgreen}{\ding{51}} & \underline{81.4} & \textbf{1743} & \underline{69.7} & \textbf{88.2} & 67.5 & \textbf{78.8} \\
	\bottomrule
\end{tabularx}
%     {Core component ablation on Qwen3-VL-8B. Every variant includes $\mathcal{D}_{origin}$; component columns indicate embodied EDA data ($\mathcal{D}_{emb}$), general VQA data ($\mathcal{D}_{gen}$), cross-entropy (CE), and GCR. }
\caption{Ablation study results on Qwen3-VL-8B-Instruct. Bold indicates top performance; underline, second-best.}
\label{tab:component_ablation}
\end{table*}

\begin{figure}[!tbp]
	\centering
	\includegraphics[width=1\columnwidth]{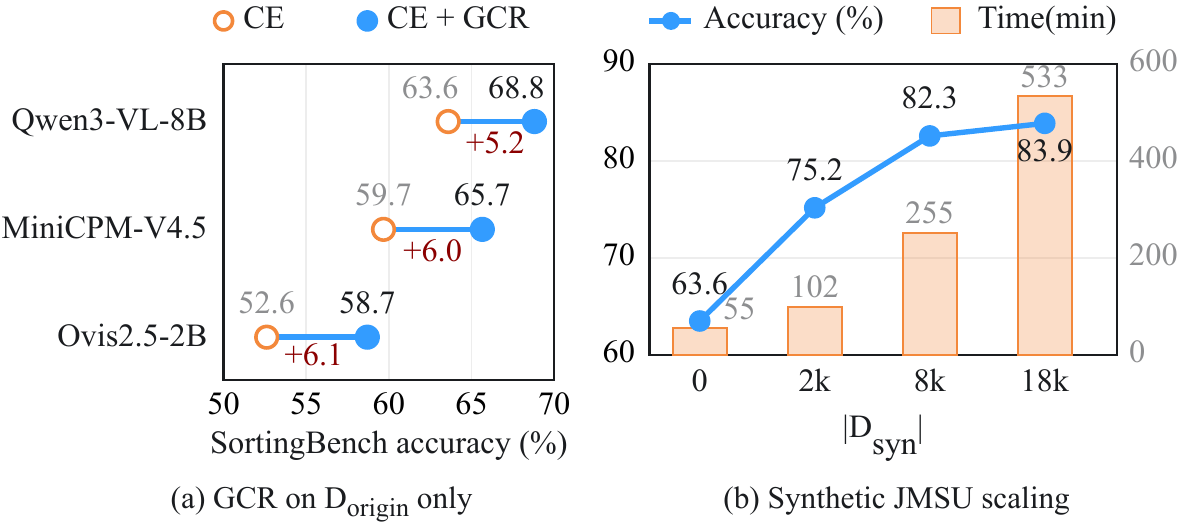}
	\caption{(a) GCR gains on $\mathcal{D}_{origin}$ only. (b) Synthetic JMSU scaling on Qwen3-VL-8B.}
	\label{fig:gcr_eda}
\end{figure}

\begin{figure*}[tbp]
\centering
\includegraphics[width=2\columnwidth]{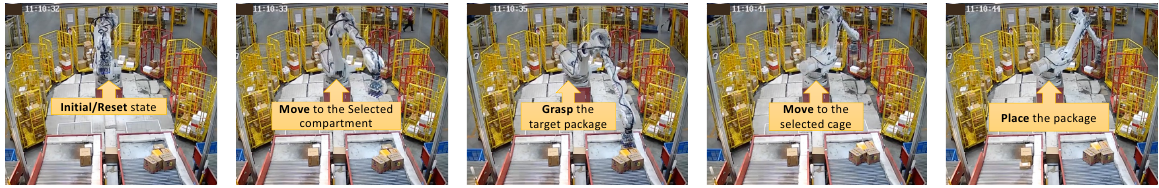}
\caption{ALSS execution: reset, move to the chosen compartment, grasp the target package, move to the chosen cage, and place.}
\label{fig:deployment}
\end{figure*}

\textbf{{Order and distractor stress tests}.}
To verify the effectiveness of \system in alleviating the problem of long visual context, we conducted stress tests by changing the images’ order and adding distractors.
Table~\ref{tab:stress} shows less than $0.3\%$ variation across independent shuffles, confirming that the randomized training policy suppresses fixed-position shortcuts. Adding $6$–$12$ irrelevant but in-domain images exposes a different failure: SFT falls to $45.8\%$ and $18.8\%$, whereas \system retains $76.8\%$ and $69.8\%$. GCR alone also degrades less than SFT. These results support the intended mechanism, \ie GCR helps the VLM preserve relevant evidence when visual context is diluted.

\textbf{{Independent effects of GCR and EDA}.}
We further verified the impact of each component.  As shown in Figure~\ref{fig:gcr_eda} (a), applying GCR improves all three models by $5.2\%$–$6.1\%$, showing its effectiveness. For EDA, Figure~\ref{fig:gcr_eda} (b) varies only the number of synthetic JMSU samples while holding SFT fixed. The first 2,000 samples provide an $11.6\%$ gain with a training time of $102$ minutes, while larger subsets continue to improve accuracy with higher cost. Therefore, the main model in Section~\ref{sec:4.2_benchmark_results} uses 2,000 synthetic JMSU samples to balance performance, training time, and the other data types.

%\textbf{{Ablation study}.}
%{Table~\ref{tab:component_ablation} restores the complete component ablation. Removing GCR from \system reduces SortingBench from $78.8%$ to $77.2%$. With $\mathcal{D}_{origin}$ alone, GCR provides the larger $63.6%\rightarrow68.8%$ gain in Table~\ref{tab:gcr_only}, indicating partial saturation when EDA already provides dense supervision. Removing embodied EDA data reduces SortingBench from $78.8%$ to $73.3%$ and BLINK/OL from $69.7%$ to $63.1%$. Removing general VQA retains $78.4%$ on SortingBench but lowers MMBench/EN from $81.4%$ to $77.3%$ and MME from $1743.0$ to $1661.0$. EDA is therefore the main source of task adaptation, while general VQA data controls forgetting and GCR improves reliance on the complete state.}

\textbf{{Ablation study}.}
Table~\ref{tab:component_ablation} presents the full component ablation.
To analyze the impact of EDA, we categorize the generated samples into embodied-related samples ($\mathcal{D}_{emb}=\mathcal{D}_{atomic}\cup \mathcal{D}_{syn}\cup \mathcal{D}_{aux}$) and general VQA samples ($\mathcal{D}_{general}$).  Removing $\mathcal{D}_{emb}$ leads to a significant performance drop of $5.5\%$ on SortingBench $(78.8\% \rightarrow 73.3\%)$, and the fundamental perceptual capabilities also decline (\eg $69.67\% \rightarrow 63.11\%$ on BLINK/OL) due to the added $\mathcal{D}_{general}$.  This confirms that training on raw data alone is insufficient for understanding complex tasks, whereas EDA provides the necessary training data density.  Conversely, excluding $\mathcal{D}_{general}$ maintains high performance on SortingBench but causes a regression in general capabilities (\eg $81.4\% \rightarrow 77.25\%$ on MMBench/EN and $1743 \rightarrow 1661$ on MME). The general data acts as a regularizer, preserving the VLM’s general capabilities during domain adaptation.
To evaluate the gain from GCR, we compare it against a standard cross-entropy loss baseline.  Using all training data, introducing GCR yields a $ 1.6\%$ improvement on SortingBench ($77.2\% \rightarrow 78.8\%$).  Additionally, using GCR on the original dataset achieved a $5.2\%$ improvement ($63.6\% \rightarrow 68.8\%$).  More importantly, on BLINK/VS, which relies on cross-image visual comparison, performance increases by $3.85\%$ ($84.3\% \rightarrow 88.15\%$).  This shows that GCR optimizes the model’s zero-shot cross-scene reasoning capabilities in distributed perception tasks, rather than simply overfitting the JMSU data distribution.

\subsection{Execution Example}
\label{sec:4.4_case_study}

Figure~\ref{fig:deployment} shows how a predicted plan is executed by the real ALSS with 2 compartments and 10 cages.  In real-world deployment testing, the \system-optimized Qwen3-VL-8B performed sorting planning for over 15,000 packages and achieved a prediction accuracy of $73.1\%$.
Although room for improvement remains, this result demonstrates the practical viability of incorporating VLM-based planning into ALSS.

%\textbf{{Limitations}.}
%{The quantitative evaluation covers one logistics application and four workstation layouts. Cross-domain gains on BLINK and MUIRBench are encouraging, but they do not validate operation in transportation, retail, or other autonomous systems. EDA also requires one-time human specification of a fact schema and domain constraints, and a new domain may require a new configuration. Finally, SortingBench isolates task-level planning from controller failures. Long-duration deployment, latency, recovery from execution errors, and safety under sensor faults remain necessary before claiming end-to-end autonomous operation.}

	% \input{5-1_discussion}
	\section{Conclusion}
\label{sec:6_conclusion}
This paper studies visual planning in ALSS, where an executable decision depends on jointly reasoning over spatially disjoint camera views. We formalize this setting as JMSU, introduce SortingBench and \system, a training framework that combines constraint-consistent EDA with GCR. EDA improves data utilization efficiency, and GCR encourages VLMs to integrate the complete visual state, alleviating attentional distraction caused by long visual contexts. 
Experiments across five VLMs consistently demonstrate its effectiveness, with Qwen3-VL-8B improving from $63.6\%$ to $78.8\%$ on SortingBench. The experiments further show that the improvement arises from stronger use of distributed evidence rather than positional shortcuts or domain memorization. Future work will extend JMSU to other autonomous domains and investigate long-horizon closed-loop operation and adaptation under changing system configurations.

	\bibliography{ref}
	
	% ===================== Appendix =====================
	% Uncomment to include the technical appendix (compiles cleanly with algorithm2e).
	% \appendix
	% \input{appendix}
	
\end{document}